\documentclass[journal,twoside,web]{ieeecolor}
\usepackage{tmi}
\usepackage[nocompress]{cite}
\usepackage{amsmath,amssymb,amsfonts}
\usepackage{algorithmic}
\usepackage{graphicx}
\usepackage{textcomp}
\usepackage{booktabs,makecell,multirow}
\def\BibTeX{{\rm B\kern-.05em{\sc i\kern-.025em  b}\kern-.08em
    T\kern-.1667em\lower.7ex\hbox{E}\kern-.125emX}}
\usepackage{hyperref}
\hypersetup{
  colorlinks=true, 
  citecolor=blue,     
  linkcolor=blue     
}
\begin{document}
\title{Flow-Guided Implicit Neural Representation for Motion-Aware Dynamic MRI Reconstruction}
\author{Baoqing Li, Yuanyuan Liu, Congcong Liu, Qingyong Zhu, Jing Cheng, Yihang Zhou, Hao Chen, Zhuo-Xu Cui, Dong Liang, \IEEEmembership{Senior Member, IEEE}                               
\thanks{ Baoqing Li and Yuanyuan Liu contributed equally to this work. Corresponding authors: Zhuo-Xu Cui and Dong Liang.}
\thanks{B. Li is with the School of Automation, Guangdong University of Technology, Guangzhou 510006, China}
\thanks{Y. Liu, C. Liu, Q. Zhu, J. Cheng, Y. Zhou, D. Liang and ZX. Cui are with the Research Center for Medical AI, Shenzhen Institute of Advanced Technology, Chinese Academy of Sciences, Shenzhen 518055, China}
\thanks{H. Chen is with the Department of Computer Science and Engineering, The Hong Kong University of Science and Technology, Hong Kong, China}
}

\maketitle

\begin{abstract}
Dynamic magnetic resonance imaging (dMRI) captures temporally-resolved anatomy but is often challenged by limited sampling and motion-induced artifacts. Conventional motion-compensated reconstructions typically rely on pre-estimated optical flow, which is inaccurate under undersampling and degrades reconstruction quality. In this work, we propose a novel implicit neural representation (INR) framework that jointly models both the dynamic image sequence and its underlying motion field. Specifically, one INR is employed to parameterize the spatiotemporal image content, while another INR represents the optical flow. The two are coupled via the optical flow equation, which serves as a physics-inspired regularization, in addition to a data consistency loss that enforces agreement with $k$-space measurements. This joint optimization enables simultaneous recovery of temporally coherent images and motion fields without requiring prior flow estimation. Experiments on dynamic cardiac MRI datasets demonstrate that the proposed method outperforms state-of-the-art motion-compensated and deep learning approaches, achieving superior reconstruction quality, accurate motion estimation, and improved temporal fidelity. These results highlight the potential of implicit joint modeling with flow-regularized constraints for advancing dMRI reconstruction.
\end{abstract}

\section{Introduction}
\label{sec:introduction}

\IEEEPARstart{D}{ynamic} Magnetic Resonance Imaging (dMRI) enables noninvasive visualization of time-resolved physiological processes in tissues and organs, offering valuable information for precise clinical diagnosis and treatment planning. However, its high spatiotemporal resolution is fundamentally limited by the slow acquisition speed of MRI, leading to a trade-off among temporal resolution, spatial resolution, and signal-to-noise ratio (SNR). To alleviate this limitation, $k$-space undersampling is commonly adopted to accelerate acquisition, but it inevitably introduces aliasing artifacts, necessitating reconstruction algorithms that can recover diagnostic-quality images from incomplete data. Among existing approaches, Compressed Sensing (CS) \cite{CSMRI} and Parallel Imaging (PI) \cite{GRAPPA} are two major paradigms. CS exploits signal sparsity priors \cite{LustigMD} to reconstruct images from undersampled measurements, typically assuming sparse representations in transform domains such as wavelet \cite{GuerquinKern,ChaariLP}, Fourier \cite{LustigMD}, or learned dictionaries \cite{CaballeroPRH}. With the evolution of reconstruction theory, single sparse priors have been extended to multi-prior models such as the low-rank plus sparse framework \cite{TremoulheacDAA,OtazoR,HuangKCCQJYZL,TingNPO}, which leverages spatiotemporal correlations to separate the dynamic sequence into low-rank background and sparse dynamic components, substantially improving reconstruction fidelity and inspiring motion-aware modeling in dMRI \cite{PanHHHKR,PanHRKH}.

Motion is inherent to dynamic imaging and, if uncorrected, can cause severe artifacts and temporal inconsistencies. Motion-compensated reconstruction seeks to address this issue by estimating and correcting motion during image reconstruction. Existing strategies primarily include deformable registration-based and optical flow-based methods. The latter directly models temporal intensity variations using the optical flow equation \cite{HornS,LucasK}, which relates spatial and temporal gradients to infer inter-frame motion fields. Representative works such as CS+M \cite{AvilesR} and MC-JPDAL \cite{ZhaoN} incorporate optical flow into CS frameworks, while DLMCR \cite{SchmodererT} and \cite{MatthiasJ} further extend optical flow modeling to sparse dictionary and complex-valued domains. However, these methods rely on finite-difference approximations of spatial derivatives, which introduce numerical errors and instability—especially under sparse temporal sampling. Although multi-scale \cite{ZhaoN} and smoothness constraints \cite{ZimmerBW} can mitigate such issues, they cannot fundamentally overcome the discretization limitations inherent in pixel-based formulations, restricting the reliability of motion estimation and reconstruction quality.

Implicit Neural Representation (INR) \cite{MildenhallSTBRN} provides a promising solution by representing signals as continuous coordinate-based functions parameterized by neural networks. Unlike traditional discrete representations, INRs enable direct computation of spatial and temporal gradients, making them particularly suitable for modeling continuous motion fields in dynamic imaging. Recent studies have demonstrated the strong capacity of INRs to capture fine spatiotemporal structures in medical imaging \cite{HuangLPCRH,Shen0FZW,CatalnT,WeiYWLC,FengJ,BaikD}, motivating the development of an INR-based framework for unified image reconstruction and motion estimation.

Building on this insight, this study proposes a motion-aware dMRI reconstruction framework that jointly models dynamic image sequences and motion fields using two coupled INRs. One INR parameterizes the spatiotemporal image content, while the other represents the optical flow. The two are coupled through the optical flow equation, which serves as a physics-inspired regularization alongside a data consistency constraint enforcing fidelity to undersampled $k$-space measurements. This formulation enables direct gradient computation in the spatiotemporal domain, fundamentally eliminating the need for finite differences and allowing simultaneous recovery of coherent images and accurate motion fields.

The main contributions of this work are summarized as follows.
\begin{itemize}
   \item A dual-INR framework is proposed to jointly model dynamic images and optical flow fields in a continuous domain. The coupling via the optical flow equation imposes physics-informed regularization and facilitates continuous gradient propagation.
   \item The method operates in a fully unsupervised manner, requiring neither external training data nor pre-estimated motion. Through the joint optimization of image representation and motion fields, the proposed approach effectively alleviates reconstruction degradation under undersampled conditions.
   \item In dynamic cardiac cine imaging, particularly at high acceleration factors, the proposed framework markedly outperforms existing INR-based and supervised deep learning methods. Benefiting from the integration of the optical flow constraint, the model precisely characterizes cardiac contraction dynamics, whereas comparative approaches tend to yield blurred reconstructions during contraction phases.
\end{itemize}

The remainder of this paper is organized as follows. Section \ref{sec:backgrounds} provides the relevant background knowledge. Section \ref{sec:methods} details the proposed methodology. Sections \ref{sec:experiments} and \ref{sec:results} present the experimental setup and corresponding results, followed by the discussion and conclusions in Sections \ref{sec:discussion} and \ref{sec:conclusion}.
%

\section{Backgrounds}
\label{sec:backgrounds}
\subsection{Dynamic MRI}
\label{sec:2.1}
The core of dynamic MRI reconstruction is to recover high-resolution image sequences from spatiotemporal undersampled k-space data. Consider a system with $N_c$ receiver coils. The dynamic image sequence is denoted as $I \in \mathbb{C}^{N_x \times N_y \times N_t}$, and the sensitivity map of each coil $c$ is $S_c \in \mathbb{C}^{N_x \times N_y \times N_t}$ $(1\leq c \leq N_c)$. The k-space data of the $c$-th coil is then given by
\begin{align}\label{eq1}
y_c = F(S_c \odot I) + n_c,  
\end{align}
where, $y_{c}$ is the undersampled k-space data collected by the $c$-th coil, $F$ is the Fourier transform (FFT) operator with under-sampled mask, which simulates the undersampling process from the image domain to the k-space, $\odot$ is the product of elements (Hadamard product), indicating the modulation of the coil sensitivity to the image, and $n_{c}$ is the noise vector that may occur during the acquisition process. The undersampled k-space data of all coils is combined to reconstruct the image sequence by minimizing the following objective function.
\begin{align}\label{eq2}
\min_{I}\frac{1}{2}\sum_{c = 1}^{N_c}\|y_c - F(S_c \odot I)\|_2^2+\lambda R(I),
\end{align}
where $\|y_c - F ( S_c\odot I)\|_2^2$ denotes the data consistency loss, which ensures that the reconstructed image remains consistent with the acquired real k-space data after undersampling to k-space. $R(I)$ represents the spatiotemporal regularization, which uses the spatial continuity and temporal smoothness of dynamic images to constrain undersampling artifacts.


Combining temporal and spatial priors, the objective equation Eq.(\ref{eq2}) can be expressed as:
\begin{align}\label{eq3}
\min_{I}\frac{1}{2}\sum_{c = 1}^{N_c}\|y_c - F(S_c\odot I)\|_2^2+\lambda_s R_s(I) + \lambda_t R_t(I),
\end{align}
where $R_s(I)$ denotes the spatial regularization term, aimed at suppressing spatial noise and artifacts, which is typically constrained using Total Variation (TV) \cite{KnollFC}. $R_t(I)$ represents the temporal regularization, which constrains smooth changes between adjacent frames. Commonly used methods include low-rank regularization and temporal sparse transform \cite{TremoulheacDAA,OtazoR}.
\subsection{Temporal Constraints with Physical Priors}
\label{sec:2.2}
Cardiac dynamic imaging requires acquiring multiple frames in a short time to capture details of cardiac motion, thus demanding high temporal resolution. While the aforementioned temporal constraints have certain effects, they lack practical physical meaning. We aim to find a physical prior in the temporal direction (t), learning the motion field during heartbeats to guide dynamic reconstruction between frames. 

The optical flow constraint is the fundamental equation describing pixel motion in image sequences, and its core idea is that the brightness of the same pixel remains constant before and after motion \cite{HornS,LucasK}. Specifically, for a point $(x, y)$ in an image, assuming its brightness at time $t$ is $I(x, y, t)$, and it moves to $(x+dx, y+dy)$ at time $t+dt$, with the brightness at this moment being $I(x+dx, y+dy, t+dt)$. According to the brightness constancy assumption:
\begin{align}\label{eq4}
I(x,y,t) = I(x+dx, y+dy, t+dt),
\end{align}
a first-order Taylor expansion of the brightness $I(x+dx, y+dy, t+dt)$ is performed around the point $(x, y, t)$.
\begin{align}\label{eq5}
\begin{split}
&I(x+dx,y+dy,t+dt)\\ 
&\approx I(x,y,t) + \frac{\partial I}{\partial x}dx + \frac{\partial I}{\partial y}dy + \frac{\partial I}{\partial t}dt,   
\end{split}
\end{align}
where, $\frac{\partial I}{\partial x} = I_x$ and $\frac{\partial I}{\partial y} = I_y$ represent the spatial gradients of the image, while $\frac{\partial I}{\partial t} = I_t$ denotes the temporal gradient, that is, the rate of change of brightness over time. We substitute the above equation into the brightness constancy assumption and simplify to obtain:
\begin{align}\label{eq6}
I_xd_x + I_yd_y + I_td_t = 0.
\end{align}
We introduce the motion velocity of pixels (optical flow vector), defined as $u = \frac{dx}{dt}, v = \frac{dy}{dt}$, and dividing both sides of the equation by $dt$ yields the optical flow constraint equation:
\begin{align}\label{eq7}
 I_xu + I_yv + I_t = 0.   
\end{align}
\section{Methods}
\label{sec:methods}
We apply the optical flow constraint equation to MRI, providing a simple physical prior modeling in the temporal direction. The temporal regularization term is defined by the optical flow constraint (Eq.(\ref{eq7})), and the objective function for MRI reconstruction in this case is as follows:
\begin{align}\label{eq8}
\begin{split}
\min_{I}&\frac{1}{2}\sum_{c = 1}^{N_c}\|y_c-F (S_c\odot I)\|_2^2 + \lambda_sTV(I)\\
&+\lambda_t \sum_{x = 1}^{N_x}\sum_{y = 1}^{N_y}\sum_{t = 1}^{N_t}\|I_xu + I_yv + I_t\|_2^2. 
\end{split}
\end{align}

We regard the cardiac motion field as the optical flow field. Since the tissue and texture of the heart do not change during its beating, the amplitude of each frame remains constant during dynamic sampling, satisfying the generalized brightness constancy assumption. In this case, $I_x$ and $I_y$ still represent the spatial gradients of the image, $I_t$ is the rate of change of amplitude over time, and u and v are the velocity vectors of the heart's movement along the x and y directions, respectively.

There are two critical issues that need to be addressed when solving the objective equation presented in Eq.(\ref{eq8}): First, the dynamic sequence $I$ in the equation is a discrete value, making it impossible to compute $I_x, I_y, I_t$ by taking its partial derivative; Second, the estimation of the optical flow vectors $u$ and $v$. Sections \ref{sec:3.1} and \ref{sec:3.2} will elaborate on our specific solutions to these problems.

\begin{figure*}[t]
\centerline{\includegraphics[width=\textwidth]{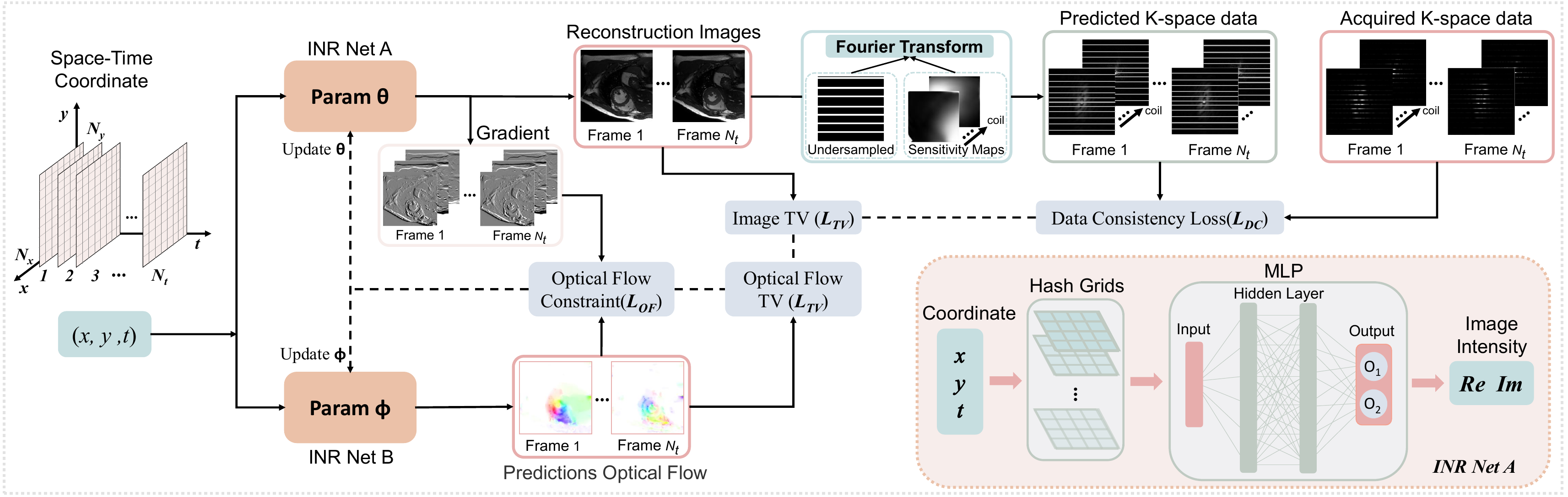}}
\caption{Framework of the proposed model. Spatiotemporal coordinates are input into two independent INR networks with identical architectures (INR Net A is shown as an example in the figure, which is processed through Hash encoding to serve as input for the MLP). The INR Net A outputs dual-channel image intensity, which is combined with real and imaginary parts to generate the reconstructed image. The reconstructed image undergoes undersampled FFT and sensitivity map modulation to obtain predicted k-space data, which is compared with real k-space data to calculate data consistency loss. The INR Net B outputs predicted optical flow vectors, which are orthogonalized with image gradients to obtain the optical flow regularization loss. Total variation is applied to both the output image sequence and optical flow vectors to constrain spatial noise. The loss function is minimized to iteratively update the parameters of the two INR networks.}
\label{fig.model1}
\end{figure*}

\subsection{INR-Based Image Reconstruction}
\label{sec:3.1}
The Implicit Neural Representations (INR) takes the spatiotemporal coordinate information of an image as input and learns the continuous mapping relationship from coordinates to pixel values (such as RGB values) through a parameterized neural network, thereby achieving super-resolution scaling of the image. Existing studies have shown that INR can be used to solve the problem of dynamic MRI reconstruction.

Specifically, we construct a continuous function $i(\theta)$ with learnable parameter $\theta$: $(x, y, t) \rightarrow \mathbb{C}$, which maps spatiotemporal coordinates to the image intensity. Our image sequence $I$ can be rewritten as $\mathcal{I}(\theta)\in \mathbb{C}^{N_x \times N_y \times N_t}$, generated entirely by INR. Our objective task then becomes finding an optimal parameter $\theta$. The objective equation as follows:
\begin{align}\label{eq9}
\begin{split}
\min_{\theta}&\frac{1}{2}\sum_{c = 1}^{N_c}\big\|y_c-F\big(S_c\odot\mathcal{I}(\theta)\big)\big\|_2^2 + \lambda_s TV\big(\mathcal{I}(\theta)\big)\\
&+\lambda_t \big\|\mathcal{I}_x(\theta)u + \mathcal{I}_y(\theta)v + \mathcal{I}_t(\theta)\big\|^2. 
\end{split}
\end{align}
Where, $\mathcal{I}_x(\theta),\mathcal{I}_y(\theta)$ and $\mathcal{I}_t(\theta)$ are all partial derivatives of the continuous function $\mathcal{I}(\theta)$, eliminating the need to compute partial derivatives concerning the discrete image $I$.
\subsection{INR-Based Optical Flow Prediction}
\label{sec:3.2}
In the optical flow constraint equation shown in Eq.(\ref{eq7}), we need to obtain the values of the optical flow vectors $u$ and $v$, which, in the context of MRI, correspond to the velocity vectors of the heart's motion along the $x$ and $y$ directions. Since the Eq.(\ref{eq7}) is underdetermined, additional constraints must be incorporated to estimate the optical flow. In the field of natural image reconstruction, gradient-based methods such as the Lucas-Kanade algorithm and Horn-Schunck algorithm are commonly employed. However, these methods rely on image grayscale variations and local information. In the MRI reconstruction domain, the acquired images are highly undersampled. Using the aforementioned methods leads to inaccurate gradient calculations, thereby affecting the accuracy of optical flow estimation.

Inspired by INR, we can also construct another continuous function $\mathcal{G}(\phi)$ to directly learn the end-to-end mapping from the spatiotemporal coordinates of the image to the optical flow field. The mapping relationship is as follows: 
\begin{align}\label{eq10}
\big(\mathcal{G}_u(x,y,t;\ \phi),\ \mathcal{G}_v(x,y,t;\ \phi)\big) = \mathcal{G}(x,y,t;\ \phi),
\end{align}
here, $\mathcal{G}(\cdot)$ is a mapping function modeled by INR, which takes spatiotemporal coordinates $(x,y,t)$ and parameter $\phi$ as inputs and outputs the two-dimensional optical flow vector $(u,v)$. INR map discrete spatiotemporal coordinates $(x,y,t)$ to a continuous optical flow field through a neural network. Without explicitly relying on fully sampled images, it can fit the optical flow values at any spatiotemporal position by optimizing only the parameter $\phi$.

\subsection{Proposed Framework}
\label{sec:3.3}
Ultimately, our dynamic MRI reconstruction problem can be formulated as solving for the optimal parameters $\theta$ and $\phi$ of two continuous functions $\mathcal{I}(\theta)$ and $\mathcal{G}(\phi)$:
\begin{align}\label{eq11}
\begin{split}
\min_{\theta,\phi}&\frac{1}{2}\sum_{c = 1}^{N_c}\big\|y_c-F\big(S_c\odot\mathcal{I}(\theta)\big)\big\|_2^2 + \lambda_{s1}TV\big(\mathcal{I}(\theta)\big) \\
&+\lambda_t \big\|\mathcal{I}_x(\theta)\mathcal{G}_u(\phi) + \mathcal{I}_y(\theta)\mathcal{G}_v(\phi) + \mathcal{I}_t(\theta)\big\|^2\\
&+ \mu \big(\big\|\nabla \mathcal{G}_u(\phi)\big\|^2 + \big\|\nabla \mathcal{G}_v (\phi)\big\|^2\big) +\lambda_{s2} TV\big(\mathcal{G}(\phi)\big). 
\end{split}
\end{align}
Here, $TV\big(\mathcal{I}(\theta)\big)$ constrains the spatial noise of the reconstructed image, $TV\big(\mathcal{G}(\phi)\big)$ constrains the spatial noise of the predicted optical flow, and $\|\nabla \mathcal{G}_u(\phi)\|^2$ and $\|\nabla \mathcal{G}_v(\phi)\|^2$ denote the smoothness constraints of the optical flow. The framework diagram of our proposed model is shown in Fig. \ref{fig.model1}.

\subsubsection{Loss Functions}
The parameters $\theta$ and $\phi$ of the two MLPs are updated using the gradient descent method, and the loss function of Eq.(\ref{eq11}) can be written in the following form.
\begin{align}\label{eq12}
\mathcal{L} = \mathcal{L}_{DC} + \mathcal{L}_{OF} + \mathcal{L}_{TV}.
\end{align}

Where, $\mathcal{L}_{DC}$ is the data consistency (DC) loss in k-space. We adopt a hybrid loss that combines relative $L_2$ loss and relative $L_1$ loss as the DC loss. This hybrid loss not only retains the advantages of relative loss, making it suitable for high-frequency dynamic range images, but also integrates the two relative errors to balance robustness and prediction accuracy. Specifically, we denote the predicted value $F\big(S_c\odot\mathcal{I}(\theta)\big)$ of the $c$-th coil as $\hat{y}_c$, and both the predicted value and measured value are written in tensor form, then the DC loss is:
\begin{align}\label{eq13}
\mathcal{L}_{DC} = \frac{\|\hat{Y} - Y\|_2}{\|Y\|_2 + \epsilon} + \frac{\|\hat{Y} - Y\|_1}{\|Y\|_1 + \epsilon}.
\end{align}
Where $\hat{Y} = [\hat{y}_1, \hat{y}_2,...,\hat{y}_{N_c}]$ is the predicted k-space data, and $Y = [y_1, y_2,...,y_{N_c}]$ is the measured undersampled k-space data. $\epsilon$ = $10^{-4}$ is a small value that prevents the denominator from becoming zero when the measured value is close to zero, ensuring the stability of numerical calculations.

$\mathcal{L}_{OF}$ is the optical flow (OF) regularization loss, which consists of the optical flow orthogonal constraint equation and the optical flow smoothing term:
\begin{align}\label{eq14}
\begin{split}
\mathcal{L}_{OF} &= \lambda_t \big\|\mathcal{I}_x(\theta)\mathcal{G}_u(\phi) + \mathcal{I}_y(\theta)\mathcal{G}_v(\phi) + \mathcal{I}_t(\theta)\big\|^2\\
&+ \mu \big(\big\|\nabla \mathcal{G}_u(\phi)\big\|^2 + \big\|\nabla \mathcal{G}_v (\phi)\big\|^2\big) 
\end{split}
\end{align}

$\mathcal{L}_{TV}$ is the Total Variation (TV) Loss, which includes the reconstructed image TV and predicted optical flow TV:
\begin{align}\label{eq15}
\mathcal{L}_{TV} = \lambda_{s1}TV\big(\mathcal{I}(\theta)\big) + \lambda_{s2}TV\big(\mathcal{G}(\phi)\big). 
\end{align}
We update the parameters $\theta$ and $\phi$ of the two INR networks through backpropagation to minimize the model total loss $\mathcal{L}$. 

\subsubsection{Network Design}
The INR-based image reconstruction and optical flow prediction model employs the MLP with a preprocessing step. This preprocessing step maps input coordinates to a higher-dimensional space, thereby enhancing the performance of high-frequency fitting \cite{TancikSMFRSRBN}. Research \cite{MullerESK,FengJ} has shown that using hash encoding as a mapping function can significantly accelerate convergence time and reduce memory consumption. The hash grid is set to 24 levels, with each level outputting 2 features, and its size is configured as 24 to ensure high-precision output. The base resolution and per-level scale factor are set to 16 and 2.0, respectively. The hash-encoded coordinate information is input into a miniature MLP, which consists of only two hidden layers with 128 neurons in each layer, followed by a ReLU activation function. The output layer has two channels. In the image reconstruction model, these channels represent the real and imaginary parts of the MRI image, while in the optical flow prediction model, they denote the optical flow vectors $u$ and $v$ in two directions. The parameters $\theta$ and $\phi$ of the two INR networks are optimized simultaneously using the Adam optimizer \cite{KingmaB}, with the learning rate set to $0.01$ and other parameters configured as $\beta_1=0.9$, $\beta_2=0.999$ and $\epsilon=1e^{-8}$. Upon reaching the predefined number of iterations, grid coordinates are fed into their respective INR networks, yielding the reconstructed MRI and corresponding motion fields.

\section{Experiments}
\label{sec:experiments}
In this section, we elaborate on the design of the experiments. The entire process was conducted on a workstation running the Ubuntu 22.04.5 LTS (64-bit) operating system, equipped with an Intel Xeon Gold 6242R CUP (@3.10 GHz), 1.5 TB RAM, and multiple NVIDIA GeForce RTX 3090 GPU (24 GB memory each). The network was implemented using PyTorch 2.5.1 and tiny-cuda-nn.

\subsection{Datasets Description}
\label{sec:4.1}
The fully sampled cardiac cine data were collected from 29 healthy volunteers on a 3T scanner (MAGNETOM Trio, Siemens Healthcare, Erlangen, Germany) with a multichannel receiver coil array (20 coils). All in vivo experiments were conducted with IRB approval and informed consent. For each subject, 10 to 13 short-axis slices were imaged with the retrospective electrocardiogram (ECG)-gated segmented bSSFP sequence during breath holding. A total of 386 slices were collected. The following sequence parameters were used: FOV = $330 \times 330$ mm, acquisition matrix = $256 \times 256$ , slice thickness = 6 mm, TR/TE = 3.0 ms/1.5 ms. The acquired temporal resolution was 40.0 ms, and each data point had approximately 25 phases that covered the entire cardiac cycle. we applied data augmentation with stride and cropping. We slid a box of dimensions $192 \times 192 \times 18\:(x \times y \times t)$ in the dynamic images along the x, y, and t directions, and the stride along the three directions was 25, 25, and 7. Finally, 118 cases of 2D-t cardiac MR data with a size of $192 \times 192 \times 18$ were selected for testing. The coil sensitivity maps were calculated for multicoil data from the undersampled time-averaged k-space center $( 48 \times 48 )$ using the ESPIRiT \cite{ESPIRiT} algorithm.

We used two different sampling methods, Cartesian and Non-Cartesian, to simulate undersampling. For Cartesian sampling, we performed fully sampled frequency encoding (along $k_x$) and undersampled phase encoding (along $k_y$) for each frame. Both random Cartesian \cite{Cartesian} masks and variable-density incoherent spatial-temporal acquisition (VISTA) \cite{VISTA} masks were used in this work to demonstrate generalization to different masks for our method. Furthermore, the 2D golden-angle radial acquisition scheme \cite{FengLi} was employed to simulate the undersampling pattern, so as to demonstrate the performance of our method under non-Cartesian sampling. The simulation process involved transforming the image domain to the frequency domain using multi-coil NUFFT \cite{NUFFT} with golden-angle trajectories based on Fibonacci numbers. Three different acceleration rate (AF) were simulated for evaluation, including 13, 8 and 5 spokes per frame (AF=14.8, 24, 38.4).
\subsection{Baselines}
\label{sec:4.2}
In this work, we select L+S \cite{OtazoR}, Time-Depend DIP (TD-DIP) \cite{YooJGYSU}, HashINR-tTVLR \cite{FengJ}, and model-based deep learning (MoDL) \cite{AggarwalMJ} as baseline methods for comparison. L+S method is an MRI reconstruction method based on compressed sensing, which decomposes the dynamic image into low-rank (L) and sparse (S) components. TD-DIP is an unsupervised method that introduces a deep image prior and combines a low-dimensional manifold and a mapping network to reconstruct dynamic MRI. HashINR-tTVLR is a HashINR-based unsupervised dynamic MRI reconstruction algorithm that uses sparsity and low-rank regularization for constraints. MoDL is a model-based image reconstruction framework that incorporates a convolutional neural network (CNN)-based regularization prior, and the model is trained using the remaining 800 views of the dataset.


\subsection{Evaluation Metrics}
\label{sec:4.3}
We adopt Peak Signal-to-Noise Ratio (PSNR) and Structural Similarity Index (SSIM) as quantitative evaluation metrics, and use error maps between reconstructed images and ground truth to visually observe the reconstruction effect. The frame-by-frame calculation formulas for PSNR and SSIM are as follows:
\begin{align}\label{eq16}
\text{PSNR} = 10\times \log_{10}\big(\frac{1}{\text{MSE}(x,y)}\big)
\end{align}
\begin{align}\label{eq17}
\text{SSIM} = \frac{(2\mu_x\mu_y + C_1)(2\sigma_{xy} + C_2)}{(\mu_x^2 + \mu_y^2 + C_1)(\sigma_x^2 + \sigma_y^2 + C_2)}
\end{align}
Here, $x$ and $y$ represent the ground truth image and reconstructed image normalized to the range $[0, 1]$, respectively. $\text{MSE}(x,y)$ is the mean squared error of the corresponding pixel values of two images. $\mu_x$ and $\mu_y$ denote the local means of images $x$ and $y$, $\sigma_x^2$ and $\sigma_y^2$ are the variances of $x$ and $y$, and $\sigma_{xy}$ is the covariance. $C_1 = (K_1 \cdot L)^2$ and $C_2 = (K_2 \cdot L)^2$ are stability constants introduced to avoid division by zero, with default values $K_1 = 0.01$, $K_2 = 0.03$, and $L$ is the dynamic range of image intensity, set to 1 after normalization.
\subsection{Parameter Configuration}
\label{sec:4.4}
Our model tunes the hyperparameters of several regularization terms, using a grid search to select the optimal parameters. For the optical flow regularization term, $\lambda_t$ is searched over the range from $[1, 10^4]$, and $\mu$ from $[10^{-2}, 10^2]$. For TV loss, both $\lambda_{s1}$ and $\lambda_{s2}$ are grid-searched within the range $[10^{-2}, 10^2]$. To ensure fair comparison, the baseline method HashINR-tTVLR adopts the same Hash grid and MLP architecture as our method, and its hyperparameter tuning strictly follows the parameter ranges provided in the original paper.
\subsection{Experiment Setting}
We conducted multiple sets of experiments to evaluate the performance of the model and performed ablation studies.
\begin{figure*}[t!]
\centerline{\includegraphics[width=\textwidth]{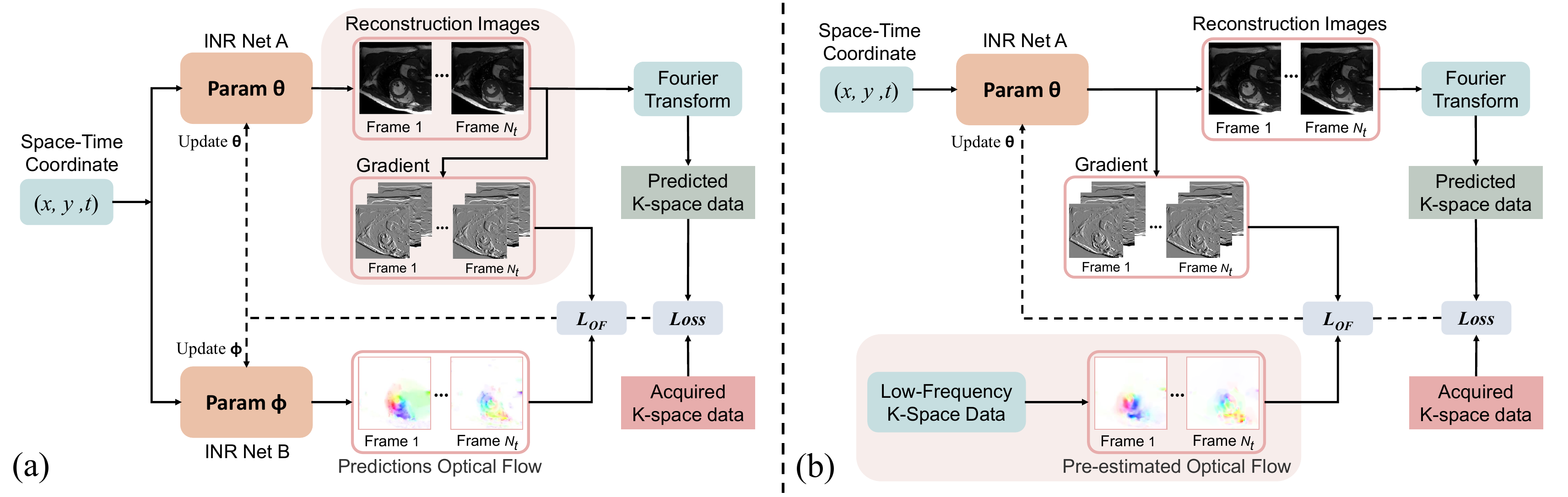}}
\caption{Frameworks of ablation study. (a) compute the gradient directly on the reconstructed MRI without utilizing the continuity of INR, and (b) pre-estimate the optical flow using low-frequency signals in the K-space (fully sampled 32×32 center region) without joint reconstruction.}
\label{fig.model2}
\end{figure*}
\subsubsection{Reconstruction Performance Comparison}
We first simulated undersampling on the retrospective dataset using Cartesian sampling. Two types of undersampling mask were employed to validate our method, which was then compared with other baseline methods. Additionally, a simulated undersampling pattern based on the 2D golden-angle radial sampling scheme was adopted to evaluate the reconstruction performance of our method under non-Cartesian sampling. The undersampling pattern and AF are shown in the Table \ref{tab.mask}.
\begin{table}[t]
  \centering
  \caption{The Undersampling Pattern and AF.}
  \begin{tabular}{lll} 
    \toprule 
    Undersampling Pattern & AF \\
    \midrule 
      Random Cartesian     & 8$\times$, 10$\times$, 12$\times$  \\
      VISTA                & 8$\times$, 10$\times$, 12$\times$, 16$\times$, 20$\times$, 24$\times$  \\
      Golden-Angle Radial  & 14.8$\times$, 24$\times$, 38.4$\times$  \\
    \bottomrule 
  \end{tabular}
  \label{tab.mask}
\end{table}
\subsubsection{Temporal Interpolation Performance Comparison}
To evaluate the temporal interpolation performance, we used the HashINR-tTVLR method as the baseline. Specifically, we designed a specific undersampling mask with full sampling in the spatial dimensions ($x$ and $y$ directions) and uniform $3$-fold undersampling in the temporal dimension ($t$ direction). This mask was used to simulate the undersampling process and the INR network was optimized accordingly. After network optimization, full-sampling spatiotemporal coordinates were input to perform INR temporal interpolation. Furthermore, we estimated a true optical flow (GT-OF) using fully sampled ground truth (GT), which was subsequently employed to guide the INR for temporal interpolation.
\subsubsection{Motion Estimation Performance}
To evaluate the accuracy of the motion field, we designed a set of simulation experiments. We took the first frame of the GT images and the corresponding motion field which was the first frame of GT-OF and generated the second frame image through inverse optical flow mapping. Subsequently using the second frame image obtained from the previous step and the corresponding motion field, we generated the third frame image using the same method. This process was repeated sequentially to obtain a set of 18-frame simulated images which we named Cheat-GT. Cheat-GT images were converted to k-space using FFT and the VISTA mask was used to simulate the undersampling process. For the Cheat-GT dataset, GT-OF was the true motion field and could serve as optical flow labels. Using this as a benchmark we evaluated the accuracy of the motion field estimated by our model during joint reconstruction.
\subsubsection{Ablation Study}
First, we conducted module ablation experiments. Specifically, we used the VISTA mask on a retrospective cardiac cine dataset to simulate undersampling with AF=12 and analyze the necessity of each constraint in the method. Second, we designed comparative experiments based on gradient calculation via image differencing to verify the superiority of the proposed INR-based continuous representation. Furthermore, we pre-estimated optical flow using ACS data and integrated it into our model to validate the advantages of the joint reconstruction method over the optical flow a pre-estimation approach. The models of the two comparative methods are illustrated in the Fig. \ref{fig.model2}.

\section{Results}
\label{sec:results}
\subsection{Reconstruction Performance}

\begin{figure*}[htbp]
\centerline{\includegraphics[width=\textwidth]{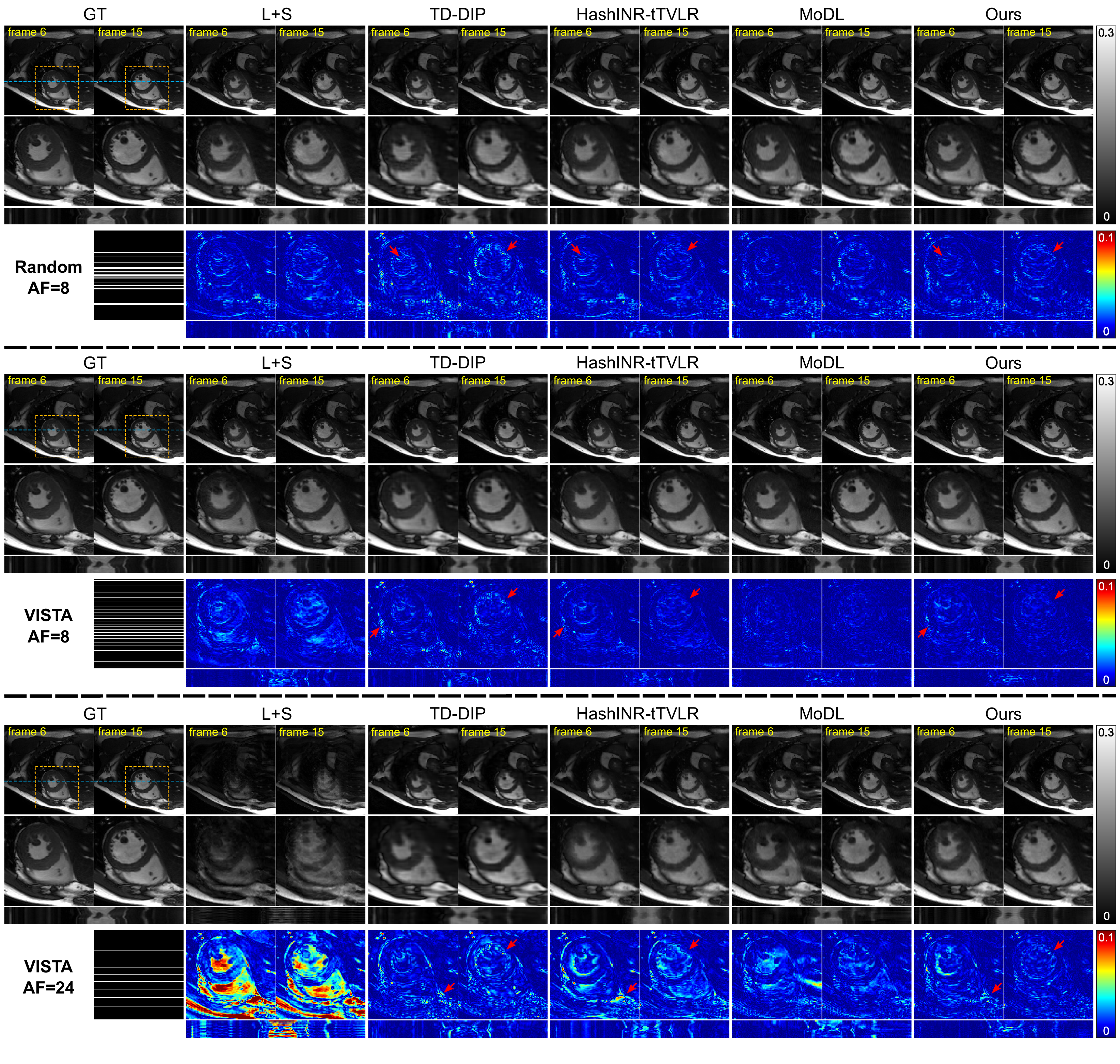}}
\caption{Reconstruction results of the proposed method and baseline methods. Visualization of reconstruction results at Random Cartesian (AF=8), and VISTA (AF=8, 24) undersampling masks. The second row shows the magnified region of the heart (orange box). The third row displays the x-t image outlined by the blue line (the 120nd slice). The fourth and fifth rows respectively show the error maps. The red arrow highlights the advantages of our method. }
\label{fig.recon2}
\end{figure*}

\begin{table*}[t]
  \centering
  \caption{Quantitative Comparison of Reconstruction Performance among Various Methods on the Cardiac Cine Dataset Is Conducted. Three Different Undersampling Patterns Are Employed: Random Cartesian (AF=8, 10, 12), VISTA (AF=8, 10, 12, 16, 20, 24), and Golden-Angle Radial (AF=14.8, 24, 38.4). The Mean Values of PSNR and SSIM Across All Test Samples, Along with Their Corresponding Standard Deviations, Are Reported. Best Results Are Shown in Bold, and Next Best Result Are Underlined.}
  \begin{tabular}{llcccccc} 
    \toprule 
    \multicolumn{2}{l}{Undersampling Pattern} & PSNR(dB) & SSIM($\times10^{-2}$)  & PSNR(dB) & SSIM($\times10^{-2}$)  & PSNR(dB) & SSIM($\times10^{-2}$) \\
    \midrule 
    \multirow{6.5}{*}{Random Cartesian} 
    & Methods\:/\:AF & \multicolumn{2}{c}{8$\times$} & \multicolumn{2}{c}{10$\times$} & \multicolumn{2}{c}{12$\times$}\\
    \cmidrule(r){2-2} 
    \cmidrule(r){3-4} 
    \cmidrule(r){5-6} 
    \cmidrule(r){7-8}
      & L+S            & 41.515 ± 1.642  & 96.484 ± 1.017  & 39.968 ± 1.858  & 95.831 ± 1.091  & 39.187 ± 1.688  & 95.157 ± 1.178 \\
      & TD-DIP         & 40.034 ± 1.616  & 94.254 ± 1.083  & 39.223 ± 1.460  & 93.431 ± 1.174  & 38.926 ± 1.550  & 92.802 ± 1.271 \\
      & HashINR-tTVLR  & \textbf{42.120 ± 1.340}  & \underline{96.870 ± 0.842}  & \textbf{41.152 ± 1.351}  & \underline{96.217 ± 0.854}  & \underline{40.483 ± 1.396}  & \underline{95.733 ± 0.950} \\
      & MoDL          & 41.175 ± 1.490 & 96.482 ± 0.867   & 39.858 ± 1.915  & 96.032 ± 0.963 & 38.781 ± 1.987 & 95.190 ± 1.124  \\
      & Ours          & \underline{42.047 ± 1.441}  & \textbf{96.878 ± 1.081}   & \underline{41.089 ± 1.381}  & \textbf{96.318 ± 1.167}  & \textbf{40.556 ± 1.390}  & \textbf{95.823 ± 1.152} \\
    \midrule 
    \multirow{13.5}{*}{VISTA}
    &  Methods\:/\:AF & \multicolumn{2}{c}{8$\times$} & \multicolumn{2}{c}{10$\times$} & \multicolumn{2}{c}{12$\times$}\\
    \cmidrule(r){2-2} 
    \cmidrule(r){3-4} 
    \cmidrule(r){5-6} 
    \cmidrule(r){7-8}
      & L+S        & 40.779 ± 1.866  & 96.373 ± 1.119  & 38.142 ± 2.063  & 94.829 ± 1.651  & 35.688 ± 2.156  & 92.391 ± 2.477  \\
      & TD-DIP     & 42.597 ± 1.667  & 96.441 ± 0.813  & 42.098 ± 1.597  & 96.118 ± 0.943  & 41.446 ± 1.841  & 95.672 ± 1.068  \\
      & HashINR-tTVLR    & \textbf{44.247 ± 1.728}  & 97.361 ± 0.897 & 42.905 ± 1.750  & 96.600 ± 1.031  & 41.909 ± 1.725  & 96.088 ± 1.026 \\
      & MoDL      & 44.016 ± 1.955  & \underline{97.481 ± 0.707} & \underline{43.470 ± 1.667}  & \textbf{97.341 ± 0.717}  & \textbf{42.893 ± 1.639}  & \textbf{97.123 ± 0.763}  \\
      & Ours            & \underline{44.212 ± 1.727}  & \textbf{97.492 ± 0.701} & \textbf{43.575 ± 1.775}  & \underline{97.325 ± 0.867}  & \underline{42.763 ± 1.781}  & \underline{96.893 ± 0.986} \\
    \cmidrule(r){2-8}
    &  Methods\:/\:AF & \multicolumn{2}{c}{16$\times$} & \multicolumn{2}{c}{20$\times$} & \multicolumn{2}{c}{24$\times$}\\
    \cmidrule(r){2-2} 
    \cmidrule(r){3-4} 
    \cmidrule(r){5-6} 
    \cmidrule(r){7-8}
      & L+S             & 32.112 ± 1.896  & 86.226 ± 3.528   & 28.421 ± 1.604  & 78.346 ± 3.708  & 26.337 ± 1.503  & 72.539 ± 3.833  \\
      & TD-DIP          & \underline{40.462 ± 1.736}  & 94.584 ± 1.283 & \underline{39.620 ± 1.632}  & 93.601 ± 1.614  & \underline{38.733 ± 1.637}  & \underline{92.744 ± 2.251}  \\
      & HashINR-tTVLR   & 40.361 ± 1.585  & 94.714 ± 1.240  & 39.014 ± 1.515  & \underline{93.644 ± 1.460}  & 37.033 ± 1.366  & 92.114 ± 1.629  \\
      & MoDL            & 40.214 ± 1.918  & \underline{95.762 ± 1.226} & 37.391 ± 1.624 & 93.104 ± 1.475 & 35.717 ± 1.665 & 91.955 ± 1.747  \\
      & Ours            & \textbf{41.825 ± 1.684}  & \textbf{96.334 ± 1.105} & \textbf{40.750 ± 1.604}  & \textbf{95.668 ± 1.198}  & \textbf{40.630 ± 1.623}  & \textbf{95.700 ± 1.370}  \\
    \midrule
    \multirow{6.5}{*}{\makecell[c]{Golden-Angle\\ Radial}}
    &  Methods\:/\:AF & \multicolumn{2}{c}{14.8$\times$} & \multicolumn{2}{c}{24$\times$} & \multicolumn{2}{c}{38.4$\times$}\\
    \cmidrule(r){2-2} 
    \cmidrule(r){3-4} 
    \cmidrule(r){5-6} 
    \cmidrule(r){7-8}
      & NUFFT           & 27.707 ± 0.786  & 56.754 ± 3.099 & 24.920 ± 0.722  & 45.861 ± 2.930 & 23.117 ± 0.666  & 37.302 ± 2.971\\
      & TD-DIP          & \underline{45.202 ± 1.442}  & 98.069 ± 0.442 & 43.068 ± 1.304  & 96.849 ± 0.604 & \underline{41.751 ± 1.353}  & 95.612 ± 0.783\\
      & HashINR-tTVLR   & 45.029 ± 1.369  & \textbf{98.543 ± 0.321} & \underline{43.362 ± 1.358}  & \underline{97.745 ± 0.389} & 41.301 ± 1.480  & \underline{96.778 ± 0.804}\\
      & Ours            & \textbf{45.901 ± 1.466}  & \underline{98.405 ± 0.383} & \textbf{43.909 ± 1.443}  & \textbf{97.826 ± 0.393} & \textbf{41.996 ± 1.545}  & \textbf{97.022 ± 0.467}\\
    \bottomrule 
  \end{tabular}
  \label{tab:result}
\end{table*}
Fig. \ref{fig.recon2} compares the reconstruction performance of representative data from the cardiac cine dataset by different methods under different sampling conditions. Visually, with the Random mask at 8$\times$ acceleration, each method can reconstruct relatively clear images, while our method reconstructs images that preserve better details. In contrast, the L+S and HashINR-tTVLR methods are more affected by noise. TDDIP causes the left papillary muscle to become smoothed during the diastolic phase, failing to preserve details. As can also be seen from the error maps, our method is more advantageous in denoising and preserving reconstructed details. In addition, when using the VISTA mask, the reconstruction performance of various methods is relatively satisfactory under 8$\times$ undersampling. As the AF increases, the proposed method exhibits significant advantages. Under extremely high acceleration factors(AF=24), the L+S method exhibits severe aliasing artifacts due to undersampling. The TD-DIP method loses most detailed information, with noticeable artifacts visible in the reconstructed images. For HashINR-tTVLR, reconstruction quality is particularly poor during the systolic phase, and error maps further indicate noise-induced degradation of cardiac contour integrity. The MoDL method also results in significant artifacts and noise during systole due to the poor generalization ability of the model. In contrast, our method only presents blurred motion artifacts at the end-systolic phase, with performance at all other time frames significantly surpassing that of comparative methods. 

Table \ref{tab:result} presents the quantitative evaluation metrics for the cardiac cine dataset from all subjects, with Cartesian sampling utilizing all 118 views and radial sampling utilizing 30 selected views. Under Cartesian sampling, our method achieves the highest SSIM and exhibits the best PSNR across most samples. Notably, at extremely high acceleration factors (AF$=20/24$), it delivers the highest PSNR ($40.750/40.630$) and SSIM ($0.957/0.957$), with improvements of $1.13/1.897$ in PSNR and $0.021/0.03$ in SSIM compared to other methods. And under radial sampling, our method achieves better PSNR while ensuring high SSIM.
\subsection{Temporal Interpolation Performance}
\begin{figure}[t!]
\centerline{\includegraphics[width=\columnwidth]{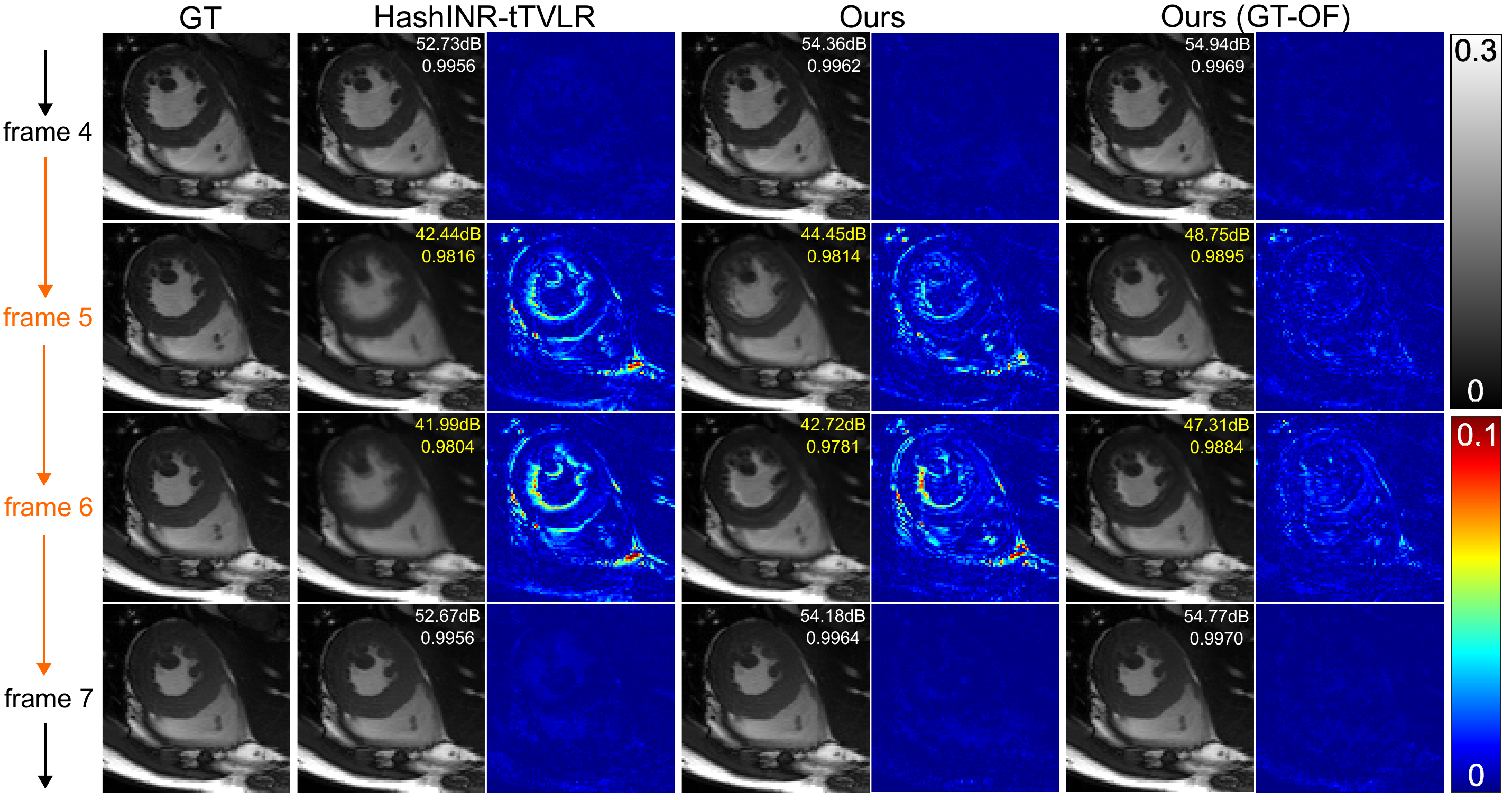}}
\caption{Temporal interpolation performance comparisons. The orange color indicates the frames generated by INR interpolation.}
\label{fig.TI}
\end{figure}

The results of temporal interpolation are shown in the Fig  \ref{fig.TI}. In 3-fold temporal frame interpolation, our model slightly outperforms the baseline method, with the interpolated frames exhibiting fewer motion artifacts. As indicated by the evaluation metrics, our model achieves better PSNR values while SSIM remains comparable. Furthermore, we use ground truth optical flow (GT-OF) to guide INR for temporal interpolation. Incorporating (GT-OF) into the model can completely eliminate blur artifacts caused by cardiac contraction, thereby reconstructing clearer and more accurate images.
\subsection{Motion Estimation Performance}
\begin{figure}[t!]
\centerline{\includegraphics[width=\columnwidth]{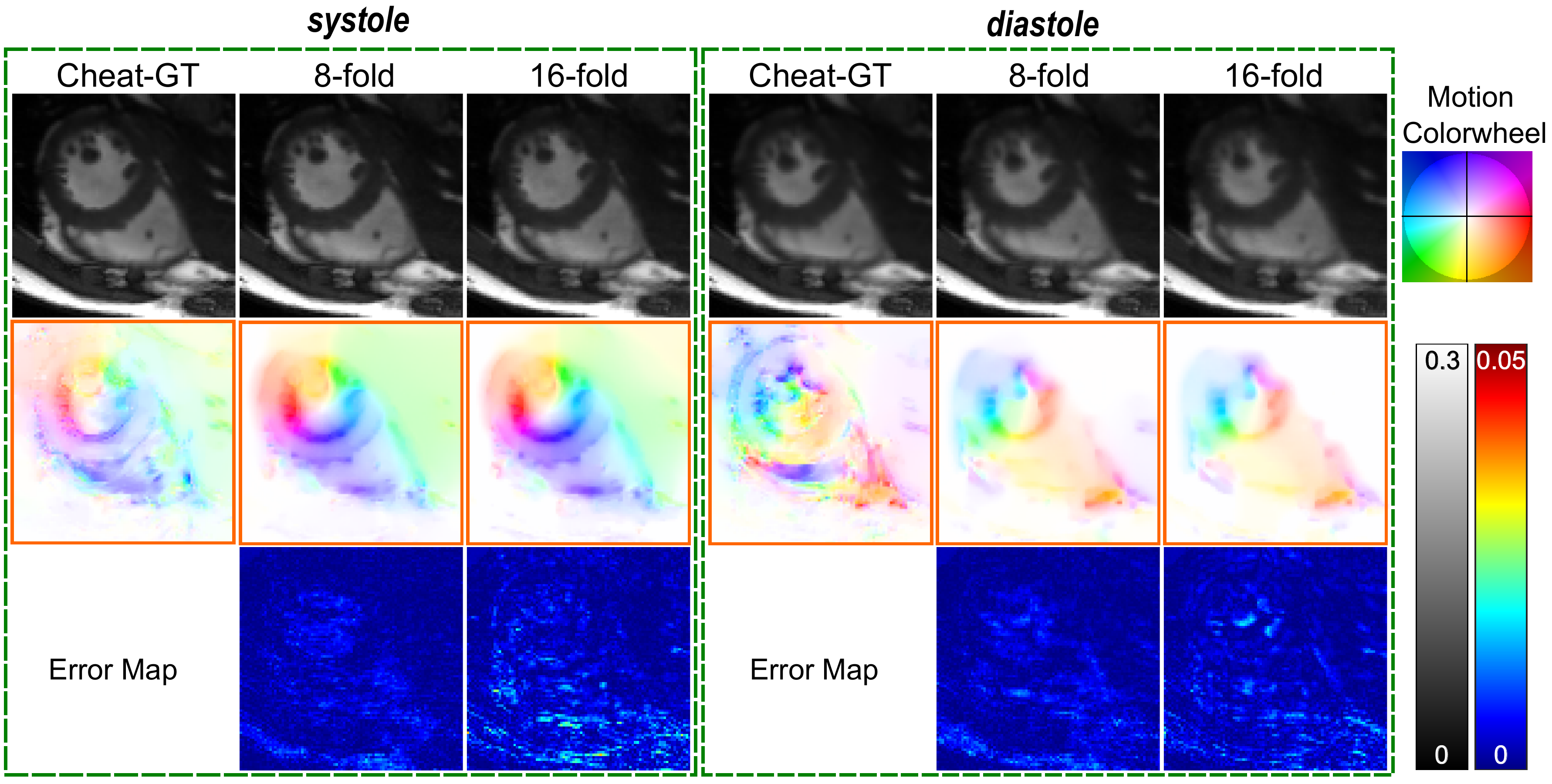}}
\caption{Reconstruction results and optical flow estimation under different AF on simulated samples (Cheat-GT). The left side represents systole, while the right side represents diastole. The first row shows Cheat-GT and the reconstruction results of different AFs. The second and third rows show the motion field and error map respectively. The top-right corner shows a color wheel representing the motion field.}
\label{fig.simi}
\end{figure}

Optical flow estimation results for systole and diastole are separately presented using the motion color wheel \cite{BakerSLRBS}, as shown in the Fig. \ref{fig.simi}. From the figure it can be observed that during systole the estimated motion fields are generally consistent. Even under extremely high acceleration magnification some motion details can still be retained. During diastole the model can accurately estimate the cardiac motion direction. At this point it is observed that motion details are not fully estimated. This is because constructing the simulated data Cheat-GT introduces estimation errors. Frame-by-frame generation further causes the later frames to become smoothed. The later frames of Cheat-GT themselves exhibit severe detail loss thus only relatively smooth motion fields can be estimated.


\subsection{Ablation Study}
\begin{figure*}[t!]
\centerline{\includegraphics[width=\textwidth]{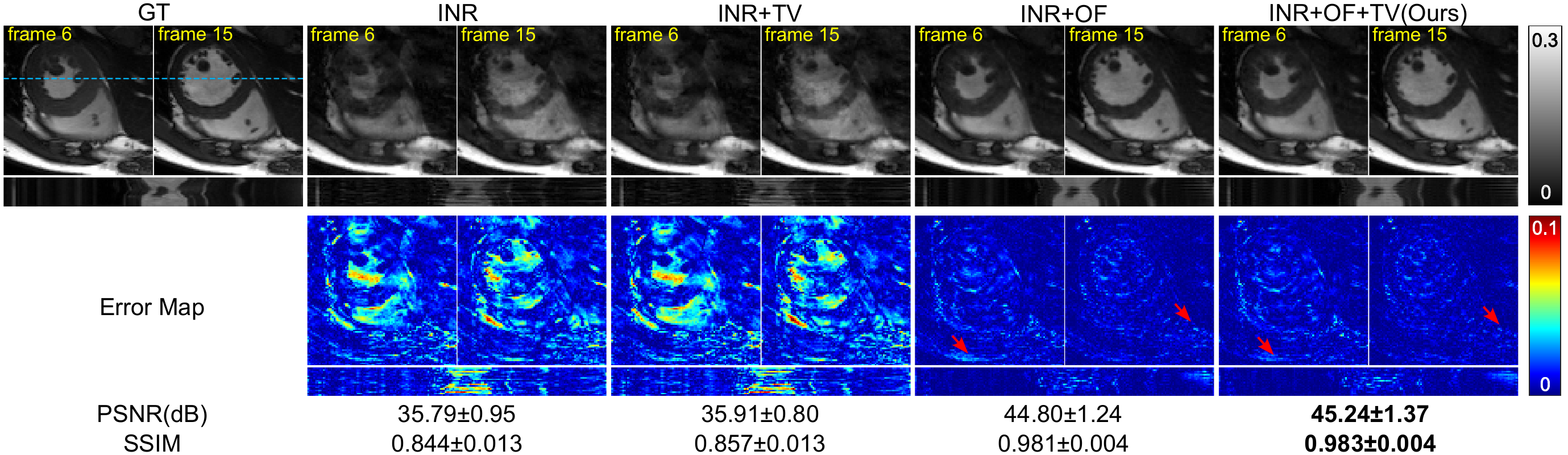}}
\caption{Results of module ablation study on cardiac cine dataset at 12-fold acceleration. The first row presents the fully sampled ground truth and reconstruction results of different methods (magnified region). The second row displays the x-t image (extraction of the 120nd slice along the x and temporal dimensions). The third and fourth rows respectively show the error maps. The red arrow points out that adding TV constraints to the optical flow regularization model can better preserve details and boundary information.}
\label{fig.module_ablation}
\end{figure*}

\begin{figure*}[t!]
\centerline{\includegraphics[width=\textwidth]{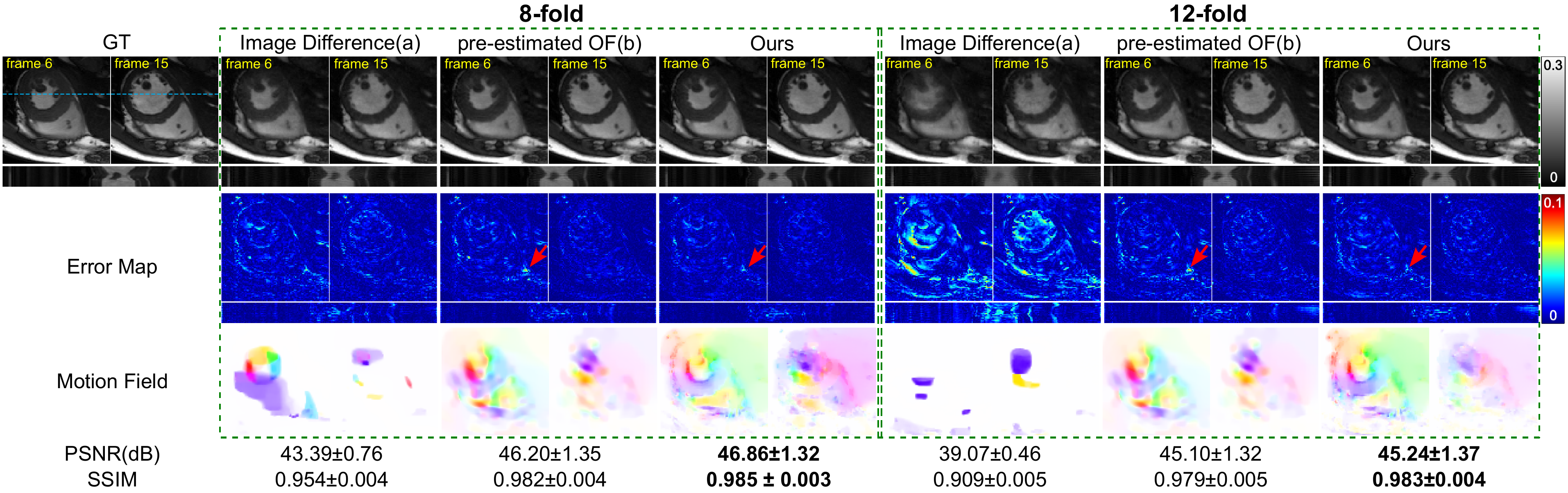}}
\caption{The ablation study results of INR continuous representation and optical flow joint reconstruction at 8-fold and 12-fold acceleration on the cardiac cine dataset. For the 8-fold reconstruction on the left, from left to right are: without INR continuous representation (a), pre-estimated optical flow from low-frequency signals (b), and our INR joint reconstruction method. The second row displays the x-t image (extraction of the 120nd slice along the x and temporal dimensions). The third and fourth rows respectively show the error maps. The last row displays the motion field. Our model can estimate more accurate motion fields and reconstruct finer high-frequency details (shown in red arrows).}
\label{fig.img_grad_pre_of}
\end{figure*}

Results of the module ablation experiments are presented in the Fig. \ref{fig.module_ablation} and Table \ref{tab.module_ablation}. As observed from the figures, optical flow-guided INR enables the reconstruction of clear temporally coherent images, and the incorporation of TV constraint more effectively suppresses spatial noise while preserving sharper boundary information (red arrows in the figure). Furthermore, quantitative evaluation metrics in the dataset further validate the necessity of each module in our method.

\begin{table}[t!]
  \centering
  \caption{Average PSNR and SSIM of the Module Ablation Study at 12-fold Acceleration on the Cardiac Cine Dataset. Best Results Are Shown in Bold.}
  \begin{tabular}{lllcc} 
    \toprule 
    INR & TV & OF & PSNR(dB) & SSIM($\times10^{-2}$) \\
    \cmidrule(r){1-3} 
    \cmidrule(r){4-5} 
     \checkmark &            &            & 34.154 ± 1.313  & 81.323 ± 2.358  \\
     \checkmark & \checkmark &            & 34.395 ± 1.515  & 83.293 ± 2.059  \\
     \checkmark &            & \checkmark & 42.320 ± 1.732  & 96.448 ± 1.088  \\
     \checkmark & \checkmark & \checkmark & \textbf{42.763 ± 1.781}  & \textbf{96.893 ± 0.986}  \\
    \bottomrule 
  \end{tabular}
  \label{tab.module_ablation}
\end{table}

Fig. \ref{fig.img_grad_pre_of} presents the experimental results of the two ablation studies, INR continuous representation and optical flow joint reconstruction. As shown in the figure, the image differencing method exhibits significant limitations. At 8-fold accelerated undersampling, only large-scale motion fields can be estimated, thereby failing to reconstruct cardiac structural details. At 12-fold acceleration, motion field estimation becomes unachievable, motion compensation methods lose efficacy, and artifacts appear in the reconstructed images. Reconstruction via pre-estimated optical flow from central low-frequency k-space signals can yield clear images, while the detail resolution is compromised (red arrows in the figure). The evaluation metrics further demonstrate that our joint reconstruction method outperforms the approach of pre-estimated optical flow.

\section{Discussion}
\label{sec:discussion}
The method proposed in this study still has certain limitations. First, at extremely high acceleration factors, blurred ring-shaped artifacts appear around the heart in some end-systolic frame images. This is due to severe undersampling, which leads to significant challenges in optical flow estimation for intermediate frames, resulting in certain deviations from the true motion state. Despite these limitations, the reconstruction results of intermediate frames still outperform those of other comparative methods, and the quantitative results at 20$\times$ and 24$\times$ accelerations in the Table \ref{tab:result} further confirm that the method exhibits significant potential in high-resolution dynamic imaging under extremely high acceleration factors. In future works, we will explore combining prior knowledge or latent mapping relationships \cite{YooJGYSU} to constrain the motion estimation of intermediate frames, reduce deviations from real motion, and capture complex motions. Furthermore, our model uses two Hash INR networks, which results in high memory usage. In the future, we will consider optimizing the network structure to reduce memory overhead. Finally, this study conducted evaluations based on a retrospective cardiac cine dataset. In future work, we will further evaluate the model's performance on prospective data and conduct more tests in different dynamic imaging scenarios.

\section{Conclusion}
\label{sec:conclusion}
This paper presented a motion-aware dMRI reconstruction framework based on INR. The model employed two INR networks to parameterize the spatiotemporal image content and the underlying motion field, respectively. Without relying on external data or prior optical flow estimation, it simultaneously reconstructed high-resolution dynamic images and estimated reliable motion vectors. Experiments on cardiac cine datasets demonstrated that the proposed method outperformed state-of-the-art motion-compensated and deep learning approaches in terms of reconstruction quality, motion estimation accuracy, and temporal fidelity. These findings underscored the potential of implicit joint modeling with motion-regularized constraints to advance dynamic MRI reconstruction.

\end{document}